\documentclass[11pt]{article}

\usepackage{acl}
\usepackage[most]{tcolorbox}
\tcbuselibrary{skins}
\usepackage{xcolor}
\usepackage{courier}
\usepackage[table]{xcolor}

\usepackage[T1]{fontenc}
\usepackage[utf8]{inputenc}
\usepackage{latexsym}
\usepackage{microtype}
\usepackage{inconsolata}
\usepackage{graphicx}
\usepackage{amsmath,amssymb}
\usepackage{newtxtext,newtxmath}

\usepackage{multirow}
\usepackage{makecell}
\usepackage{array}
\usepackage{float}
\usepackage{pifont}    % ding symbols

\newcommand{\bluecell}[1]{\textbf{\textcolor{blue}{#1}}}
\newcommand{\greencell}[1]{\textbf{\textcolor{green!50!black}{#1}}}
\newcommand{\textbfc}[1]{\textbf{#1}}
\newcommand{\cmark}{\textcolor{green!60!black}{\ding{51}}} % check
\newcommand{\xmark}{\textcolor{red!75!black}{\ding{55}}}   % cross

\tcbset{
  promptbox/.style={
    enhanced,
    colback=gray!10,
    colframe=black!50,
    fontupper=\ttfamily\small,
    boxrule=0.4pt,
    sharp corners,
    left=6pt, right=6pt, top=4pt, bottom=4pt,
    boxsep=2pt,
    before skip=10pt, after skip=10pt,
    coltitle=black,
    attach boxed title to top left={xshift=2mm,yshift=-2mm},
    boxed title style={colback=gray!20, colframe=black!50, boxrule=0.4pt, sharp corners, font=\bfseries\small},
  }
}

\title{GRAD: Generative Retrieval-Aligned Demonstration Sampler for Efficient Few-Shot Reasoning}

 \author{Oussama Gabouj \thanks{Equal contribution.} \hspace{1cm} Kamel Charaf \footnotemark[1] \hspace{1cm} Ivan Zakazov \footnotemark[1] \\ {\bf Nicolas Baldwin} \hspace{1cm} {\bf Robert West} \\
        EPFL, Lausanne, Switzerland \\
        \texttt{oussama.gabouj@gmail.com} \\
        \texttt{publication.charaf@gmail.com} \\
        \texttt{\{ivan.zakazov, nicolas.baldwin, robert.west@epfl.ch\}} \\
        }

\begin{document}

\maketitle
\begin{abstract}

Large Language Models (LLMs) achieve strong performance across diverse tasks, but their effectiveness often depends on the quality of the provided context. Retrieval-Augmented Generation (RAG) enriches prompts with external information, but its reliance on static databases constrains adaptability and can result in irrelevant demonstrations. In this work, we propose a Generative Retrieval-Aligned Demonstrator (GRAD), a dynamic demonstration-based approach where an LLM model is trained to generate input-specific concise demonstrations. By tailoring demonstrations to each input, our method offers better contextual support than traditional RAG approaches. We demonstrate the superiority of GRAD under budget constraints, where we limit both the number of tokens used per demonstration and the number of tokens used for the final output. Trained solely on a math dataset, GRAD consistently outperforms strong baselines on Qwen2.5-14B across mathematical reasoning and advanced STEM questions, highlighting GRAD’s robust generalization to out-of-distribution (OOD) domains such as physics, chemistry, and computer science. Furthermore, we show that demonstrations generated by trained smaller models can effectively guide larger target models, reducing training costs while maintaining competitive accuracy. Overall, this work introduces a scalable demonstration generator model presenting the first step toward a dynamic few-shot learning paradigm in resource-constrained settings. We release the code used for the project:
\url{https://github.com/charafkamel/GRAD-demonstration-sampler}

\end{abstract}
\vspace*{-0.3\baselineskip}
\begin{figure*}[h]
    \centering
    \includegraphics[width=1\linewidth]{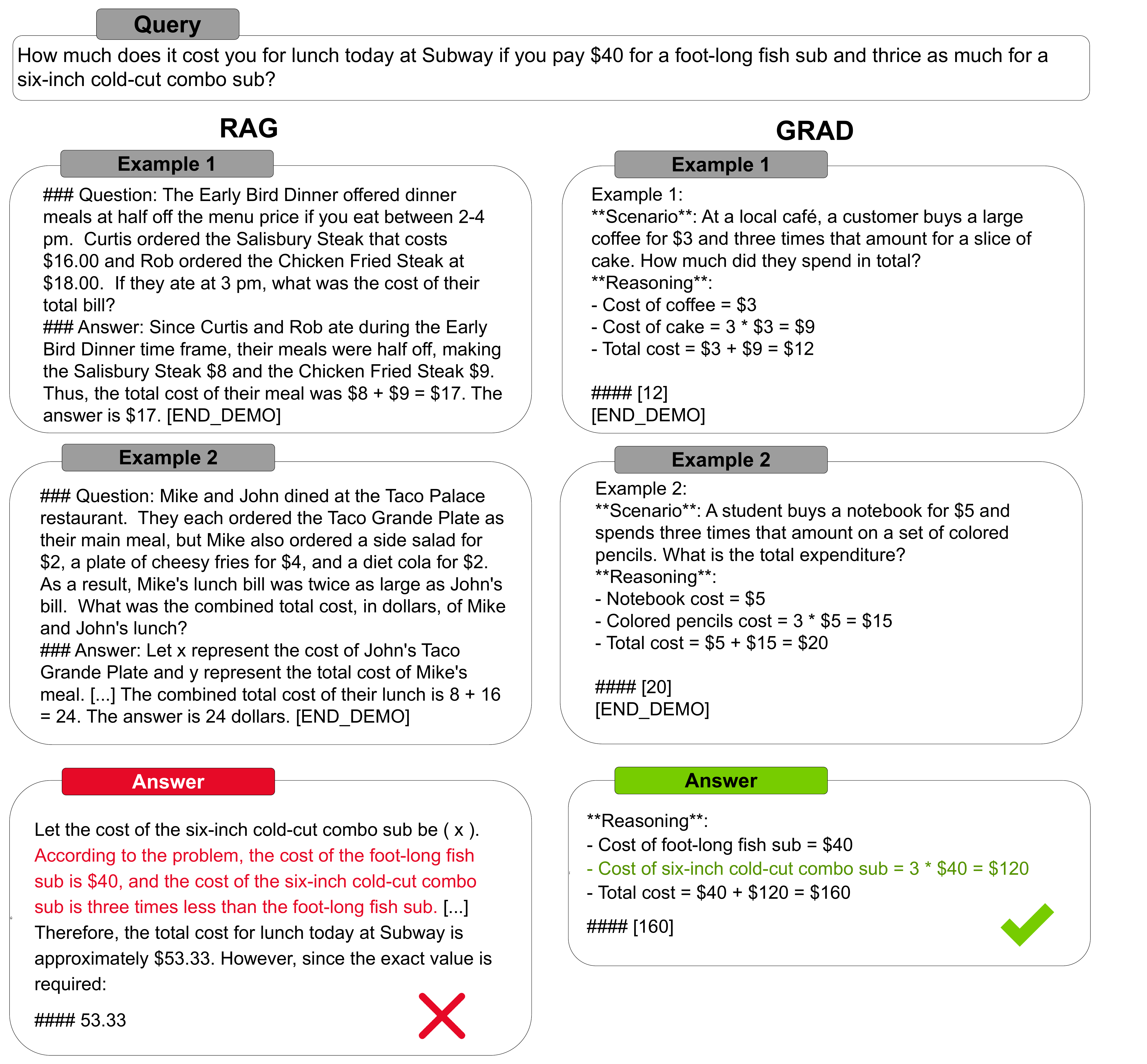}
    \caption{Example input query from GSM8K with demonstrations and outputs from RAG and GRAD. RAG retrieves demonstrations from a static database, whereas GRAD generates task-specific demonstrations within a token budget. A final output length constraint is applied in both cases; GRAD produces shorter demonstrations and a more concise final answer.}
    \label{fig:example}
\end{figure*}

\section{Introduction}

Large Language Models (LLMs) excel in multi-step reasoning tasks, often operating in a zero-shot setting. Techniques such as context augmentation and few-shot learning \cite{brown2020languagemodelsfewshotlearners} are powerful and widely used to further enhance efficiency on downstream tasks. However, the effectiveness of these methods strongly depends on the quality and relevance of the demonstrations. A well-established method to address this challenge and to enable contextualization is Retrieval-Augmented Generation (RAG) \cite{lewis2020retrieval}, \cite{guu2020retrieval}, which incorporates external knowledge by pulling information from static databases. However, due to the limited document set, retrieved examples may not always align with the query. Recent studies have also questioned the effectiveness of few-shot prompting for out-of-distribution (OOD) tasks, suggesting that when demonstrations are mismatched, they can degrade performance \cite{singhal2022assessing}. 

To address these challenges, we introduce GRAD: a Generative Retrieval-Aligned Demonstrator designed to combine the adaptability of generation with the efficiency of retrieval. GRAD dynamically generates task-specific demonstrations under a strict token budget, enabling better performance in both in-distribution (ID) and OOD tasks. To achieve this, we train GRAD with reinforcement learning (RL), which is the core mechanism to produce demonstrations, steering target models toward correct and concise generations.

Additionally, we consider a variant called GRAD\textbf{i} (GRAD \textbf{i}nitialized), where the generator is first initialized with supervised fine-tuning (SFT). This auxiliary step provides initial guidance to the demonstrator, after which the training process is identical to the one used with GRAD. Unless stated otherwise, we use \textit{GRAD} to denote our method in general, and we will explicitly distinguish between GRAD and GRADi in the results section.

We evaluate our approach on diverse reasoning benchmarks (see Section \ref{sec:evaluation_dataset}) using models ranging from 3B to 14B parameters and observe that GRAD variants consistently outperform RAG across all evaluated datasets for our largest model.

Our main contributions are as follows:
\begin{itemize}
\setlength\itemsep{-2pt}
    \item We propose GRAD, an RL-trained generative model that is optimized to produce task-specific, token-constrained demonstrations that generalize to OOD tasks.
    \item We present GRADi, an alternative to GRAD that is warm-started with SFT before applying the same RL pipeline as used in GRAD.
    \item We introduce a composite reward function that trains GRAD and GRADi to generate informative yet budget-constrained demonstrations. This effectively mitigates the tendency of large models to produce excessively long outputs, resulting in shorter, more compact answers.

\end{itemize}

\section{Related Work} \label{sec:related_work}

\paragraph{\textbf{In-context Learning and Prompting.}} 
Few-shot prompting enables LLMs to reason more effectively by embedding relevant examples into the input. \citet{brown2020languagemodelsfewshotlearners} demonstrated that prompting with examples allows large models to generalise without gradient updates. \citet{wei2023larger} showed that larger models show new capabilities in ICL settings, particularly in adapting to unusual input-output pairs that differ from those seen during training.  Subsequent work by \citet{min2022rethinking} and \citet{GoodExamplesGPT3} emphasized the importance of input format and semantic similarity between prompts and queries. \citet{kojima2022large} showed that even minimal prompts like ``Let's think step by step'' significantly change reasoning style. Complementing these findings, \citet{chen2023many} examined the effect of demonstration quantity, revealing diminishing returns beyond a small number of examples.  These studies confirm that prompting structure influences not only model accuracy but also the verbosity of the output. Building on this, our method focuses on applying ICL by generating high-quality demonstrations, enhancing the model’s ability to generate concise reasoning leading to the correct solution.

\paragraph{\textbf{RAG.}} RAG improves LLM performance by enhancing the model’s input with information retrieved from external sources. The original RAG framework laid the foundation by combining a retriever and a generator for open-domain tasks. More recently, \citet{chen2024benchmarking} systematically assessed how different LLMs benefit from RAG, and \citet{gao2023retrieval} offered a survey that outlined its challenges and design space.  Notably, while ICL traditionally relied on fixed demonstrations, integrating RAG to fetch relevant in-context examples has proven to be more effective.  \citet{liu2021makes} found that selecting semantically similar in-context examples enhances GPT-3's performance, while \citet{huang2023fewer} introduced CoT-Influx, a method that prunes less informative tokens to include more concise Chain-of-Thought examples, significantly boosting mathematical reasoning.  Despite its advantages, RAG faces challenges in OOD scenarios. \citet{finlayson2025post} highlighted that fine-tuning LLMs with RAG can lead to performance degradation when the training data is OOD, causing misalignment between retrieved content and target responses.  Motivated by these limitations, we propose a dynamic alternative: instead of retrieving, our model learns to actively generate demonstrations relevant to the input. %This approach aims to mitigate the challenges associated with RAG in OOD settings.

\paragraph{\textbf{Generative Demonstration Learning.}}   A growing line of research focuses on generating demonstrations dynamically, rather than solely relying on retrieval. Methods like Self-ICL \citet{wang2022self} and Auto-CoT \citet{zhang2022automatic} synthesize task-specific exemplars, guiding models with tailored chains of thought that often enhance performance, especially in OOD settings. More recent approaches, such as Auto-Demo Prompting \citet{feng2024auto}, create demonstrations during batch inference by reusing earlier outputs, reducing token overhead while preserving accuracy. Similarly, in long-context QA, context recycling can generate effective few-shot examples from the input passage itself \citet{cattan2024can}. These generative strategies offer fine-grained control over demonstration length and structure, balancing reasoning quality with efficiency in both token usage and compute cost. However, these methods rely on vanilla models prompted to generate demonstrations. In contrast, we propose training a model to query-adaptive, token-budgeted demonstrations that condition the target model’s reasoning.

\paragraph{\textbf{Reinforcement Learning RL.}} RL allows us to move beyond static fine-tuning by optimizing the model based on outcome-driven feedback. \citet{schulman2017proximal} introduced Proximal Policy Optimization (PPO), offering stable policy optimization. \citet{ouyang2022training} demonstrated the effectiveness of RLHF for instruction-following models, while \citet{rafailov2023direct} proposed DPO to align outputs with preferences without explicit rewards. For mathematical reasoning, recent works such as \citet{zhang2025grpo} and \citet{shao2024deepseekmath} introduced Group Relative Policy Optimization (GRPO), which promotes more stable training and improves performance on complex multi-step problems. Recent work also explores training efficiency: \citet{wang2025reinforcement} showed that minimal supervision can drive reasoning gains, and \citet{li2025adaptive} proposed AGPO to stabilize and optimize training. These findings clearly show that integrating RL in our model is particularly promising for generating highly contextually relevant demonstrations effectively.

\paragraph{\textbf{Sentence Embedding and Dataset Similarity.}} Accurately measuring semantic similarity is essential for selecting relevant demonstrations in retrieval-based systems. Motivated by this, \citet{jiang2023scaling} investigates the use of large language models for sentence embeddings, showing that in-context learning improves embedding quality without fine-tuning. \citet{zhang2024simple} proposed prompt engineering techniques like Pretended Chain of Thought and Knowledge Enhancement.  Comprehensive surveys by \citet{farouk2019measuring} and \citet{cohere2025similarity} provide insights into various approaches for measuring sentence similarity. These key insights motivate our use of similarity scoring in our work: we employ it to efficiently retrieve relevant documents in our RAG pipeline and systematically evaluate and rank datasets based on their similarity to the training data.

\section{Methodology} \label{sec:methodology}

\subsection{Data Preprocessing}
\paragraph{\textbf{Dataset Collection and Splitting.}}
The main dataset that we use is the Math Reasoning Dataset with Diverse Difficulty (MRD3) \cite{FewerIsMore}. MRD3 was created by merging the training datasets of math reasoning benchmarks - including GSM8K, MAWPS, MAWPS-single and 1000 random samples from AQuA. GPT-4 is then used to generate formatted Chain-Of-Thought (CoT) reasoning steps for each question. %ensuring consistency across examples.

The original MRD3 contains 9.7k question-answer pairs. To remove redundancy, we drop duplicated questions based on pairwise cosine-similarity over the TF-IDF \cite{salton1988term} vectors of the input queries. For each sample, we compare the similarity with all subsequent inputs and remove any entry with a similarity score of 1. After filtering, the resulting 8081 distinct samples are split into two stages. First, 10\% is set aside as an independent test set. The remaining 7273 samples are further split into evaluation (10\%), RAG (25\%) and train (65\%) subsets. The RAG corpus is used to retrieve relevant demonstrations during GRADi training and evaluation. 
During SFT training, RAG demonstrations serve as examples, showing how queries similar to the user's should be structured.
Besides training, RAG also serves as a solid baseline to compare our method with.

\paragraph{\textbf{Preparation for RAG Integration.}} \label{PreparationSFT} To integrate MRD3 into the RAG system, each data sample was transformed into a standardized structure, expressed as \texttt{Question + CoT Reasoning + Answer}. We consider one instance of this structure as a demonstration. After formatting the dataset with the new schema, each demonstration is embedded using the pre-trained sentence-transformer model \textit{all-mpnet-base-v2} \cite{MPNetTokenizer} and stored in \textit{Chroma DB} \cite{ChromaDB}. During inference, for each input query, the system retrieves the top two most relevant demonstrations from the retrieval database. These demonstrations are concatenated with the initial query and passed to a second, frozen LLM (target model), which generates the final answer (see Figure \ref{fig:train}). 

\noindent\subsection{Answer Extraction and Evaluation} \label{answer_extraction}
To ensure consistent answer extraction, we prompt the model to generate structured output by concluding its reasoning with the format \texttt{\#\#\#\# \{final answer\}}, as specified in Section~\ref{generationPompt}. In some cases, particularly with smaller models, the output may deviate from this format. To address this, we apply a regular expression designed to extract all digits, fractions, or numerical values from the model's output, retaining the last extracted number as the model's final prediction. We also track and report instances where the extraction process fails. The final answer is validated by comparing it with the expected output, with a tolerance of $10^{-4}$ for numerical values to qualify as correct.

\subsection{Token Budget and Demonstration Constraints}
To ensure a fair comparison between GRAD and our other baselines, we constrain the number of tokens used for generating both the instructions and the final output. 

\paragraph{\textbf{Instruction Length Constraint.}} \label{instructions_constraint}
As GRAD is a generative model that produces a variable number of tokens per demonstration, it is crucial to limit its output tokens to ensure fair comparison with the baseline methods. To establish a suitable token budget, we analyzed the token length distribution of the RAG demonstrations in the RAG split and found that, on average, each RAG sample has 150 tokens. As we retrieve 2 demonstrations from RAG for each input query, to match this number with our GRAD-generated demonstrations, we had to apply a cap of 300 tokens, i.e. demonstrations are truncated if they exceed this threshold.
As this limit is also applied during RL training, the model will learn to generate demonstrations that are short and complete under this constraint. Having this constraint, we also eliminate the possibility that GRAD outperforms RAG due to more generated tokens provided as a context.

\paragraph{\textbf{Final Output Length Constraint.}} \label{final_output_constraint}
To ensure comparability with the RAG baseline and to prevent unnecessary token usage, we limit the final output of the model (i.e. the reasoning trace and the final answer) to 256 tokens. We find that the average length of the final output in the MRD3 dataset is around 150 tokens, demonstrating that effective reasoning can be expressed concisely - further supporting the 256-token limit as a practical, performance-aligned design choice. Furthermore, our training encourages the model to internalize that generating shorter, high-quality demonstrations within a fixed budget improves downstream performance.

\subsection{Training}
\begin{figure*}[h]
    \centering
    \includegraphics[width=1\linewidth]{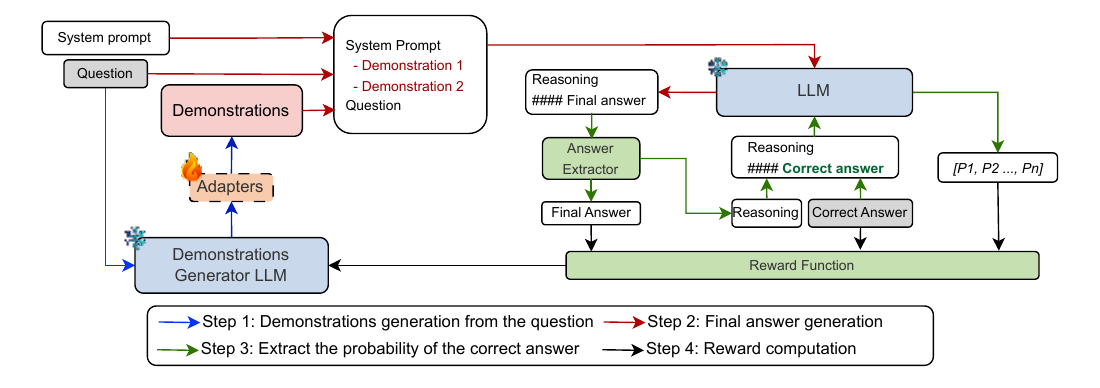}
    \caption{Overview of the GRAD pipeline. Step 1: The model generates demonstrations, which are concatenated with the system prompt and the user's query to form the context. Step 2: The context guides the target model to produce a reasoning trace and the final answer. Step 3: The predicted answer is replaced with the correct answer and passed through the frozen LLM to compute the token-level log probabilities. Step 4: Computing a multi-objective reward to ensure confidence and correctness of the final answer and compliance with the token budget.}
     \label{fig:train}
\end{figure*}

\subsubsection{Supervised Finetuing (SFT)}
For the SFT-only baseline and for the warm-up stage of GRADi,  we train the LLM to replicate the behavior of the RAG system. To do this, we use training data formed by pairing each input query with its corresponding retrieved demonstrations. For each input query, we ask the model to generate two relevant demonstrations as explained in Section \ref{PreparationSFT}.  This process helps the model learn to output well-formed and coherent reasoning paths. Although the SFT-aligned demonstrator is effective in generating high-quality outputs in ID settings, it often struggles with OOD scenarios, because SFT is optimized to mimic reference RAG demonstrations, rather than to explicitly maximize the final answer accuracy. As a result, the model may produce suboptimal demonstrations when faced with unfamiliar inputs. To overcome this limitation, we introduce RL to directly align demonstration generation with the final task objective, thereby enabling more robust generalization in OOD scenarios.

\subsubsection{Reinforcement Learning (RL)}
We evaluated and compared two RL algorithms, namely PPO and GRPO. While PPO is commonly used in previous works, GRPO consistently offered better stability and efficiency in our experiments. Therefore, we adopt GRPO for all final results reported in this work. Our RL training pipeline (as shown in Figure \ref{fig:train}) follows four core steps:

\paragraph{\textbf{Step 1: Demonstration Generation.}}
The input query is passed to the Demonstration Generation module, which produces a set of relevant few-shot demonstrations within the instruction length constraint defined in Section \ref{instructions_constraint}. 

\paragraph{\textbf{Step 2: Final Answer Generation.}} The generated demonstrations are combined with the original query and the system prompt, then passed to the target model. The model produces a complete reasoning process and a final answer in the same format as the provided demonstrations. The final answer is extracted from the output (as described in Section \ref{answer_extraction}), and the entire output is constrained to 256 tokens (as explained in Section \ref{final_output_constraint}).

\paragraph{\textbf{Step 3: Log Probability Extraction.}} To evaluate how well the demonstrations support correct answer generation, we replace the final answer of the output reasoning with the correct answer and feed it back into the frozen LLM. The model then computes the log probabilities of the tokens in the correct answer, denoted as [$P_1$, $P_2$, ..., $P_n$]. These log probabilities provide a fine-grained view into how confidently the model predicts each token of the correct answer given the \textit{context}, i.e. input query + generated demonstrations + target model reasoning.

\paragraph{\textbf{Step 4: Multi-objective Truncation-Aware Reward.}} \label{par:reward} The reward function is designed to improve model accuracy while encouraging the generation of useful demonstrations without exceeding a fixed token budget. Inspired by \citet{huang2023fewer}, the reward comprises three components: Log Probabilities Reward ($R_{\text{p}}$), Accuracy Reward ($R_{\text{acc}}$), and Demonstration Count Reward ($R_{\text{demo}}$).

The \textbf{$R_{\text{p}}$} evaluates the model's ability to predict the correct tokens based on their log probabilities as defined by Equation \ref{rp}. The $L_{\text{llm}}$ is the mean negative log probability of the correct answer given the \textit{context}. We use the mean log probability (i.e., geometric mean of token probabilities) to prevent bias against longer answers. By construction, lower $L_{\text{llm}}$ values (i.e., higher predicted probabilities) correspond to higher rewards. This equation not only rewards confident and correct predictions but also emphasizes the value of less confident but correct predictions. Moreover, it provides partial credit for incorrect answers when the correct option also has high confidence.

\begin{equation} \label{rp}
R_p = \frac{1}{1 + L_{\text{llm}}}
\end{equation}

The \textbf{$R_{\text{acc}}$} is designed to encourage the model to produce a correct and complete final answer. It is a binary reward: the model receives a score of 1 if the final answer is correct and is fully generated (i.e., not truncated) and 0 otherwise. This truncation-aware formulation encourages the model to prioritize accuracy while staying within the token budget limit.

The \textbf{$R_{\text{demo}}$} given by Equation \ref{rdemo} incentivizes the generation of valid demonstrations. Let $n$ denote the number of generated demonstrations and $D$ the expected target count. We set $D = 2$ to match the RAG baseline.  We maximize the  $R_{\text{demo}}$ for GRAD if it generates four demonstrations, as more concise and relevant demonstrations would typically exceed the 300-token budget. This maintains fair comparison with the RAG baseline, as we use the same token limit, and it allows for greater diversity.
\begin{equation} \label{rdemo}
R_{\text{demo}} = \frac{n}{D} \cdot \mathbb{1}_{\{n \leq 4\}}
\end{equation}

This cap ensures the model doesn't exploit the reward function by generating excessive or low-quality demonstrations. Since at least 100 tokens are allocated per demonstration, the token budget naturally limits the number of useful demonstrations, and this rule further prevents reward hacking.

The Final Reward is given by Equation \ref{rf}:
\begin{equation}
\label{rf}
\text{Reward} = R_p + R_{\text{acc}} + R_{\text{demo}}
\end{equation}

In summary, the reward balances 3 objectives: ensuring answer accuracy ($R_{\text{acc}}$), rewarding high-confidence reasoning ($R_p$), and promoting the generation of relevant, concise, and valid demonstrations ($R_{\text{demo}}$) all while respecting the token budget.

\subsection{Demonstration sampling strategies} \label{sec:pipelines}

In our framework, GRAD refers to the RL-only variant, and GRADi refers to the version that is initialized with an additional SFT stage before RL. During evaluation, we compare both variants with the baseline strategies. All the demonstration samplers are prompted to generate two factually correct and different examples using less than 300 tokens. All target models are instructed to generate the correct final answer in the pre-defined format using only 256 tokens (see \ref{generationPompt}).
The following provides a brief explanation of each strategy considered. %, including baselines and our GRAD variants.

\paragraph{Zero-shot model.} The model generates answers without demonstrations, and an extractor retrieves the final response. This zero-shot setup serves as the baseline for comparison.

\vspace{-5pt}
\paragraph{RAG.} 
The model retrieves two documents from a fixed database based on query similarity. These are combined with the input to guide the LLM’s answer. While RAG improves performance using added context, it’s limited by the fixed database, which may lack relevant examples for OOD queries.

\paragraph{SFT-only model.}
We conducted another evaluation on the models which were trained with SFT but without RL. This pipeline also serves as a baseline for comparison with the GRAD variants.

\paragraph{BASE model.}
In this setup, the untrained (vanilla) model creates its own demonstrations. These self-generated demonstrations are then used to answer the final question. Although this setup benefits from the model's ability to adapt dynamically to the input query, the lack of training might result in suboptimal performance. 

\paragraph{GRAD model.} GRAD is our RL-only variant, in which the model is optimized directly to generate task-specific demonstrations using a composite reward (see Step 4 in Section \ref{par:reward}). These demonstrations are concatenated with the input to provide richer context, allowing the model to adapt to diverse queries and overcome the limitations of RAG.

\paragraph{GRADi model.} GRADi combines the two training stages by first initializing the generator with SFT and then continuing optimization with RL. Unlike the SFT-only baseline, where no additional training is applied after the SFT step, GRADi uses SFT as a warm-start to stabilize the format and the structure of the demonstrations before using RL.

\begin{table*}[h]
    \centering
    \renewcommand{\arraystretch}{1.105}
    \begin{tabular}{l|l|*{6}{c}}
        \hline
        \multirow{3}{*}{\textbf{Model}} & \multirow{3}{*}{\textbf{Method}} & \multicolumn{6}{c}{\textbf{Datasets}} \\
        \cline{3-8}
        & & \textbf{GSM8K} & \textbf{\makecell{draw\\structured}} & \textbf{MathQA*} & \textbf{\makecell{deepmind\\basic\_math}} & \textbf{\makecell{ARC\\Challenge}} & \textbf{MMLU*} \\
        \hline

        \multirow{6}{*}{\makecell{Qwen2.5\\7B\\Instruct}} 
        & Zero-shot  & 83.89 & 36.50 & 44.79 & 67.78 & 87.71 & 62.22  \\
        & RAG        & 83.59 & 36.50 & 43.78 & 63.33 & 85.92 & 59.51  \\
        & SFT-only   & 74.05 & 41.00 & 42.38 & 60.00 & 84.39 & 59.88 \\
        & BASE       & \textbfc{85.80} & 38.50 & 49.73 & 64.44 & \textbfc{88.14} & 62.59  \\
        & \bluecell{GRAD}   & 84.27 & \bluecell{43.00} & \bluecell{54.72} & \bluecell{70.00} & 88.05 & \bluecell{64.20} \\
        & \greencell{GRADi}       & 84.73 & \greencell{47.00} & \greencell{53.11} & \greencell{68.89} & 88.05 & \greencell{62.71}  \\
        \hline

        \multirow{6}{*}{\makecell{LLaMA\\3.1-8B\\Instruct}} 
        & Zero-shot  & 78.24 & 42.00 & 44.04 & 43.33 & \textbfc{83.53} & 49.13  \\
        & RAG        & 76.79 & 29.50 & 40.67 & 38.89 & 73.89 & 39.75  \\
        & SFT-only   & 71.91 & 33.50 & 39.91 & \textbfc{60.00} & 75.51 & 42.72 \\
        & BASE       & 75.73 & 37.50 & 39.97 & 48.89 & 82.51 & \textbfc{52.47}  \\
        & \bluecell{GRAD}   & \bluecell{78.85} & \bluecell{46.50} & \bluecell{45.12} & 46.67 & 80.80 & 50.00 \\
        & \greencell{GRADi}       & 77.10 & \greencell{43.00} & 42.70 & 38.89 & 81.48 & 51.23  \\
        \hline

        \multirow{6}{*}{\makecell{Qwen2.5\\14B\\Instruct}} 
        & Zero-shot  & 72.75 & 30.50  & 27.73 & 58.89 & 91.13 & 48.27  \\
        & RAG        & 83.89 & 27.50 & 37.50 & 64.44 & 90.70 & 48.52  \\
        & SFT-only   & 83.66 & 36.50 & 42.00 & 65.56 & 74.83 & 40.74 \\
        & BASE       & 84.12 & 34.00 & 43.78 & 70.00 & \textbfc{92.32} & 59.75  \\
        & \bluecell{GRAD}    & \bluecell{90.92} & \bluecell{40.50} & \bluecell{56.98} & \bluecell{72.22} & 91.64 & \bluecell{65.31} \\
        & \greencell{GRADi}       & \greencell{90.46}  &\greencell{45.00} & \greencell{57.80} & \greencell{70.00} & 91.98 & \greencell{65.06}  \\
        \hline
    \end{tabular}
    \caption{Performance Comparison across Models and Methods (Accuracy in \%). The same backbone model is used for both the demonstration sampler and the target model. Datasets are ordered by their semantic similarity from left to right in decreasing order. Blue indicates cases where GRAD outperforms all baselines (independent of GRADi), while green indicates cases where GRADi does so (independent of GRAD). If the best-performing model on a given benchmark is neither GRAD nor GRADi, it is reported in bold.}
    \label{tab:performance_cleaned}
\end{table*}

\subsection{Evaluation Setup} \label{sec:evaluation_dataset}
\noindent\paragraph{Evaluation Datasets.}

To evaluate the model's ability to generalize beyond the training domain, we conduct experiments on ID and OOD benchmarks. Specifically, we assess our strategies on the \textit{GSM8K} dataset, which serves as the primary ID benchmark, and on five diverse OOD datasets: \textit{MMLU}, \textit{MathQA}, \textit{draw-structured}, \textit{DeepMind basic\_math}, and \textit{ARC\_challenge}. 

For the \textit{MMLU\*} benchmark \cite{mmlu}, we selected five subsets: \textit{college\_physics}, \textit{formal\_logic}, \textit{college\_computer\_science}, \textit{college\_chemistry}, and \textit{machine\_learning}, and grouped them under a single merged evaluation dataset referred to as \textit{MMLU*}. Similarly, from the \textit{MathQA} dataset \cite{mathqa}, we included subsets such as \textit{physics}, \textit{gain}, \textit{other}, \textit{general}, and \textit{geometry}, and merged them into a combined dataset labelled \textit{MathQA*}. 
We use each dataset’s standard test split and report the accuracy using a predefined answer format and token-budget.

\paragraph{Dataset Similarity Computation.}
To measure similarity between each evaluation dataset and the MRD3 training set, we compute pairwise cosine similarities using sentence embeddings. For each sample in our evaluation dataset, we identify the 2 most similar training examples based on cosine-similarity, as we would do when retrieving the RAG documents. We then average these top-2 similarity scores for each test sample, and finally compute the overall similarity by averaging across the entire evaluation dataset. This results in a single similarity score between the train and a benchmark dataset. Lower values of similarity indicate that the dataset is more OOD compared to the training data.

\section{Results and Discussion} \label{sec:results_analysis}
\subsection{Models performance}

We compare the performance of our methods on six benchmark datasets. The results can be found in Table~\ref{tab:performance_cleaned}, which presents the accuracy (\%) of various models; all of them having the same backbone. Models in the table are arranged by increasing size, from the smaller models at the top (e.g. Qwen2.5-7B) to larger ones at the bottom (e.g. Qwen2.5-14B). Within each model group, six methods are compared as described in Section \ref{sec:pipelines}. The datasets are also ordered by their semantic similarity to the training dataset, from left to right: \textit{GSM8K} (most similar) to \textit{MMLU*} (most dissimilar). This setup allows us to evaluate the generalization performance of the baseline strategies and both GRAD versions.

\begin{figure*}[h]

    \centering
    \renewcommand{\arraystretch}{1.3}
    \includegraphics[width=1\linewidth]{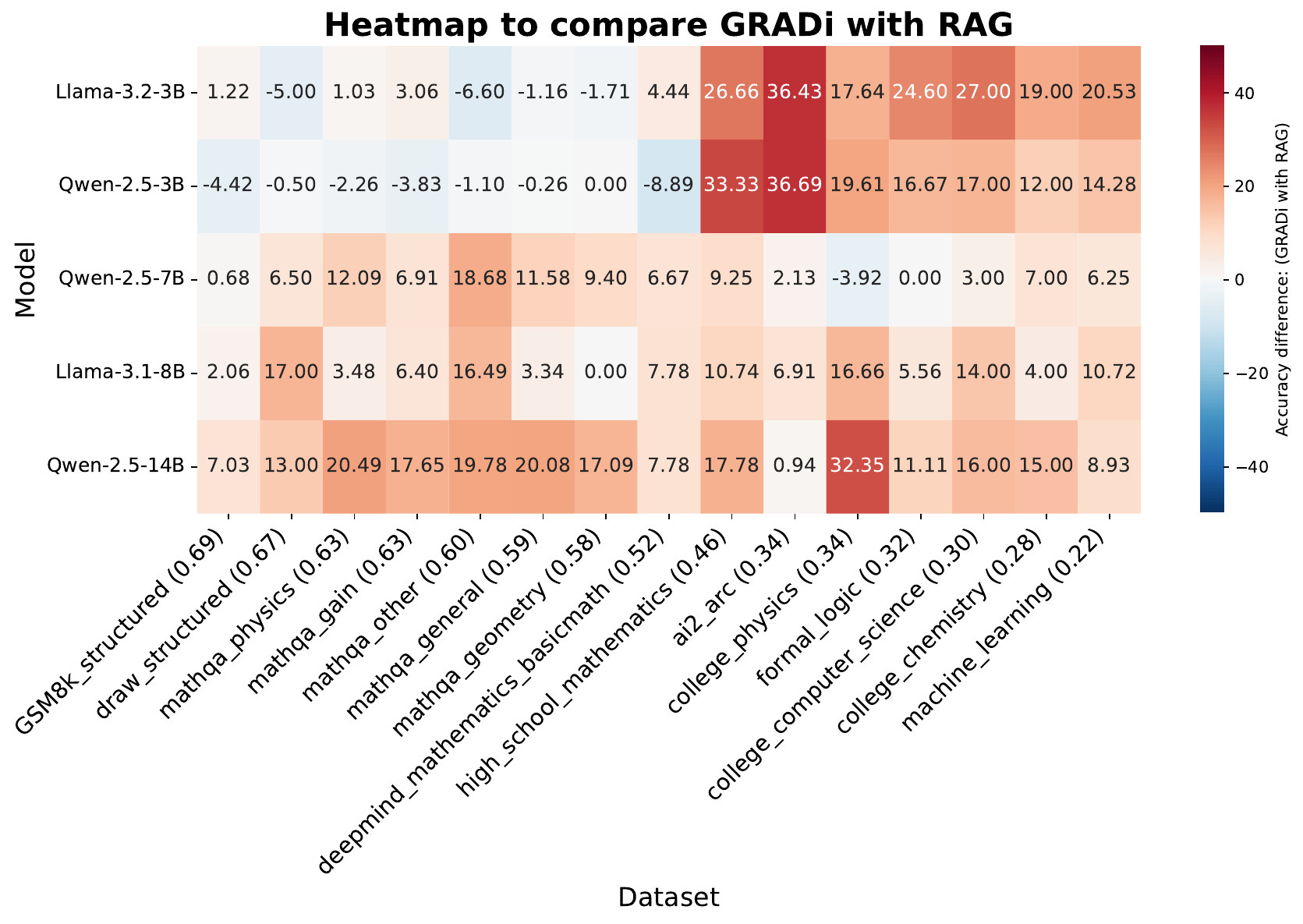}
    \caption{Heatmap of accuracy differences between GRADi and RAG. Red denotes gains for GRADi, blue for RAG, and lighter cells indicate similar performance. Datasets on the x-axis are ordered by their semantic similarity from left to right in decreasing order. Models are defined in the y-axis, ordered based on their size from top to bottom. Each cell shows the mean percentage-point difference in exact-match accuracy for that (model, dataset) pair, with the colorbar indicating magnitude and sign.}
    \label{fig:rlsft_diff}
\end{figure*}
\subsection{Results Analysis}

Table~\ref{tab:performance_cleaned} presents a consistent pattern: as model size increases, the performance of GRAD models improves substantially across all datasets. \textit{Qwen2.5-7B} achieves the highest accuracy on 4 out of 6 datasets, significantly outperforming both Zero-shot and RAG. The advantage is even more pronounced at the scale of 14B, where GRAD consistently dominates nearly all benchmarks. 

The heatmap in Figure~\ref{fig:rlsft_diff} shows how the performance of GRAD and RAG differs. Lighter shades indicate similar differences, while red and blue show gains favoring GRAD and RAG, respectively. In the top-left region, which corresponds to smaller models on more ID datasets, the differences are marginal. Moving to the right, where smaller models are evaluated on increasingly OOD datasets, GRAD shows significant improvements, demonstrating its ability to generalize beyond training distributions.  In the bottom row, the 14B model consistently outperforms RAG across nearly all datasets, highlighting the robustness of large-scale GRAD in few-shot settings. Further analysis comparing GRAD with alternative baseline approaches is presented in Appendix \ref{apx: further_analysis}

In summary, our findings confirm that dynamically generated demonstrations produced by RL-trained demonstrators outperform static retrieval, especially for OOD datasets. This advantage grows significantly with larger models, which produce better-aligned examples.

\subsection{Can Smaller Models Provide Good Demonstrations for Larger Models?}
\label{sec:strong_demos}

We investigate whether demonstrations generated by fine-tuned, smaller, cheaper models can be effectively used by a much stronger model (\textit{Qwen2.5-14B}) during inference. This approach could significantly reduce computational cost during demonstration generation. To isolate the effect of the demonstration source, we fix the answer-generation model to Qwen2.5-14B and vary only the model responsible for generating the demonstrations.

Table \ref{tab:strong_demo_eval_reverse} reports the performance when using demonstrations generated by the 3B, 7B, and 14B versions of Qwen2.5. The smaller models (3B and 7B) are fine-tuned to generate demonstrations for the bigger (14B) model, while the 14B-BASE is the vanilla, untrained version.
We observe that demonstrations generated by the smaller, fine-tuned models can yield surprisingly competitive performance when used by a much larger model for final answer generation. Fine-tuned smaller models can serve as cost-efficient demonstration generators without significantly compromising performance.

\begin{table}[H]
  \centering
  \renewcommand{\arraystretch}{1.3}

  \begin{tabular}{ >{\raggedright\arraybackslash}m{2.2cm} >{\centering\arraybackslash}m{1.3cm} >{\centering\arraybackslash}m{1.3cm} >{\centering\arraybackslash}m{1.3cm} }
    \hline
    \textbf{Dataset} & \textbf{3B-Demo} & \textbf{7B-Demo} & \textbf{14B-BASE} \\
    \hline
    GSM8K              & 66.67 & 72.22 & \textbf{84.12} \\
    draw\_structured   & 42.5  & \textbf{47.00} & 34 \\
    MathQA* & 51.23 & \textbf{54.13} & 43.78 \\
    % Use \newline for a manual line break within a cell
    deepmind\newline basic\_math & 66.67 & \textbf{72.22} & 70 \\
    MMLU* & 50.73 & 57.288 & \textbf{59.57} \\
    \hline
    
  \end{tabular}
  \caption{Accuracy (\%) of Qwen2.5-14B using demonstrations generated by itself (untrained, vanilla model) or by smaller, fine-tuned models. Datasets are ordered by their semantic similarity from top to bottom in decreasing order. The model with the highest performance on the corresponding benchmark is highlighted in bold.}
  \label{tab:strong_demo_eval_reverse}
\end{table}

\subsection{Generalization to Non-Mathematical Tasks}
While our training focused primarily on mathematical reasoning, we evaluated performance on two diverse benchmarks: ARC Challenge v2 and selected MMLU subsets (e.g., formal logic, computer science, physics). These represent a new multiple-choice setting that differs substantially from mathematical reasoning and also span distinct domains.  This evaluation allows us to assess whether the capabilities learned in a math-centered training pipeline can transfer effectively to broader reasoning challenges.  Despite this shift, on MMLU, our GRAD pipeline shows strong generalization, with larger models (e.g., Qwen2.5-14B) consistently outperforming the zero-shot baseline by more than 17.5\% and surpassing RAG by over 16.5\%. In the case of ARC Challenge v2, the zero-shot already attains high accuracy, and adding demonstrations yields no measurable gains and can introduce noise.
\newpage
\section{Conclusion} \label{sec:conclusion} 

This work introduced generative retrieval-aligned demonstration samplers (GRAD and GRADi) for creating input-aligned demonstrations under strict token budgets, that consistently outperforms traditional RAG and zero-shot methods, particularly for OOD scenarios. Our findings highlight the benefits of dynamic generation over static retrieval, particularly the generalizability of our approach beyond the training domain to OOD queries that differ in both task and domain. We also show that our approach is modular and cost-efficient: demonstrations generated by smaller, less expensive models can boost the accuracy of larger models. We also emphasize that our approach does not replace RAG. Retrieval is still applicable for knowledge-intensive tasks that might require grounding from an external corpus. Both GRAD variants can be best viewed as an alternative that aims to generalize to OOD inputs, which complements, rather than replaces, retrieval-based methods.

\section{Future work}
An important and promising direction for future research is to explore a new, hybrid GRAD (called H-GRAD) architecture that would combine the strengths of generative and retrieval-based methods. In this setup, the model would dynamically select between the top retrieved and generated demonstrations, based on the relevance to the input query. H-GRAD could maintain the robustness and the OOD generalization of GRAD while leveraging the factual advantage of RAG. We believe this line of research has the potential to deliver a more reliable few-shot reasoning system that will improve applicability and accuracy.
Additionally, we will explore how the number of demonstrations generated, the demonstration length and the token budget affect model training dynamics and the final output token length, providing deeper insights into the interplay between the number of demonstrations, demonstration size, output conciseness, and answer correctness.

\section{Limitations}

\subsection{Token Length Constraint}
In all experiments, we limited the output tokens of the final answer generation to a fixed number of 256 tokens. Complex reasoning tasks (especially those involving multiple steps) may require longer outputs to fully express the solution. We also applied a separate cap of 300 tokens for the total length of generated demonstrations i.e the model can generate as many demonstrations as it can, as long as the total remains within this token limit. While we fixed these hyperparameters for consistency and comparability, this constraint implicitly limits the number of demonstrations that can be provided, especially for tasks that require detailed examples.

\subsection{Fixed Number of Demonstrations}
Across the RAG and  GRAD pipelines, a fixed number of demonstrations were used per input query, regardless of its complexity. GRAD was trained to generate exactly two demonstrations, while RAG retrieved a static set of two examples. This constraint may lead to under-informing complex queries or overwhelming simple ones. A fixed number of examples reduces flexibility and may degrade performance on tasks with highly variable difficulty.

\subsection{Demonstration Factuality and Reliability}
While GRAD dynamically generates demonstrations tailored to each input, there is no verification step to ensure these demonstrations are factually accurate and reliable. The system assumes that the generated examples are educationally sound, but no formal check is applied during training or inference. This limitation opens the possibility for the model to use misleading reasoning paths, which could negatively impact the final answer quality. Although prior work has shown that even partially incorrect demonstrations can sometimes enhance performance, it would be valuable to filter out incorrect demonstrations. 

\section{Ethical consideration}
In this work, we demonstrate that GRAD outperforms traditional RAG methods in OOD scenarios under a budget constraint. However, generative approaches raise important ethical considerations. While RAG systems retrieve documents from a controlled database,  generative models produce demonstrations dynamically, with no inherent control over the content. GRAD-generated demonstrations might reflect some biases present in the training data or introduce false, misleading demonstrations. Since generated demonstrations are created dynamically, it is difficult to fully control and remove potentially harmful generations.

\bibliography{custom}
\appendix
\section{Appendix}
\label{sec:appendix}

\subsection{Training setup} Given a fixed computational budget, we carefully selected the parameters and techniques we used to ensure the best possible performance under resource constraints. All the obtained results in this study were computed using NVIDIA A100 GPUs with 80 GB of RAM. The input length was fixed at 256 tokens, the instruction length at 768 tokens, and the output length at 256 tokens.

\paragraph{\textbf{SFT.}} Each training session lasted less than 1 hour, using 2 epochs with a learning rate of 4e-4 and a batch size of 32 samples.

\paragraph{\textbf{RL.}} The training required significantly more computational resources due to the inclusion of multiple models and the on-the-fly generation of demonstrations. We used distributed training with 4 NVIDIA A100 GPUs. 
Initially, training took approximately 15 hours for a single epoch, using a learning rate of 1e-5 and 2 demonstrations per sample for the GRPO model. Increasing the number of demonstrations resulted in out-of-memory (OOM) errors, which is a serious limitation that had to be dealt with.

To mitigate the memory constraints, we employed Low-Rank Adaptation (LoRA), an efficient fine-tuning approach that allowed us to load only the base model and apply lightweight adapters dynamically at runtime. The same model was used for both demonstration and final answer generation. To further reduce memory consumption, we enabled gradient checkpointing, which allowed us to trade compute for memory efficiency.
Additionally, all models were loaded using 16-bit floating point precision for efficiency.

To acquire faster training speed, we employed a vllm server - an efficient, high-throughput, low-latency inference engine for LLM that introduces PagedAttention, a memory-optimized attention mechanism \cite{kwon2023vllm}. 

Using these techniques, we reduced training time to approximately 12 hours. Leveraging these strategies, we managed to train with 7 demonstrations per sample for all models, except for the 14B model, which was limited to 3 due to memory constraints.

\subsection{Used Models}
Table \ref{tab:model_summary} provides an overview of all models used in our experiments, including both LLaMA and Qwen families across varying model sizes.
\begin{table}[h]
  \centering
  \begin{tabular}{llc}
    \hline
    \textbf{Family} & \textbf{Model Name} & \textbf{Size} \\
    \hline
    Meta & Llama-3.2-3B-Instruct  & 3B \\
    Qwen & Qwen2.5-3B-Instruct       & 3B \\
    Qwen & Qwen2.5-7B-Instruct       & 7B \\
    Meta & Llama-3.1-8B-Instruct  & 8B \\
    Qwen & Qwen2.5-14B-Instruct      & 14B \\
    \hline
  \end{tabular}
  \caption{Overview of the models used in training}
  \label{tab:model_summary}
\end{table}

\subsection{Dataset similarity}
To better understand the relationship between our training data and evaluation benchmarks, we compute dataset similarity with the training dataset. Table~\ref{tab:dataset_similarity} presents these similarity scores, ordered from highest to lowest. As shown, \textit{GSM8k\_structured} is the most similar to MRD3, while \textit{machine\_learning} is the least similar. 

\begin{table}[H]
  \centering
  \renewcommand{\arraystretch}{1.1}
  \resizebox{0.48\textwidth}{!}{%
  \begin{tabular}{ >{\raggedright\arraybackslash}m{5cm} >{\centering\arraybackslash}m{1.75cm} >{\centering\arraybackslash}m{1.6cm}}
    \hline
    \textbf{Dataset} & \textbf{Similarity} & \textbf{\# samples} \\
    \hline
    GSM8k\_structured & 0.6903 & 1310\\
    draw\_structured & 0.6729 & 200\\
    mathqa\_physics & 0.6348 & 488\\
    mathqa\_gain & 0.6329 & 391\\
    mathqa\_other & 0.5981 & 91\\
    mathqa\_general & 0.5934 & 777\\
    mathqa\_geometry & 0.5839 & 117\\
    mathematics\_basicmath & 0.5167 & 90\\
    mmlu-high\_school\_mathematics & 0.4645 & 270\\
    ai2-ARC-Challenge & 0.3406 & 1172\\
    college\_physics & 0.3374 & 102 \\
    formal\_logic & 0.3216 & 126\\
    college\_computer\_science & 0.2993 & 100\\
    college\_chemistry & 0.2829 & 100\\
    machine\_learning & 0.2195 & 112\\
    \hline
  \end{tabular}}
  \caption{Semantic similarity scores between MRD3 and various evaluation datasets}
  \label{tab:dataset_similarity}
\end{table}

\subsection{Token distribution length}
In this section, we present the token length distributions for both types of outputs. The first corresponds to the output of the demonstration sampler, which reflects the instruction token length. We truncate this output at 300 tokens. The second corresponds to the output of the target model, which we truncate at 256 tokens. 
\onecolumn

\begin{figure*}[h]
    \centering
    \includegraphics[width=0.8\linewidth]{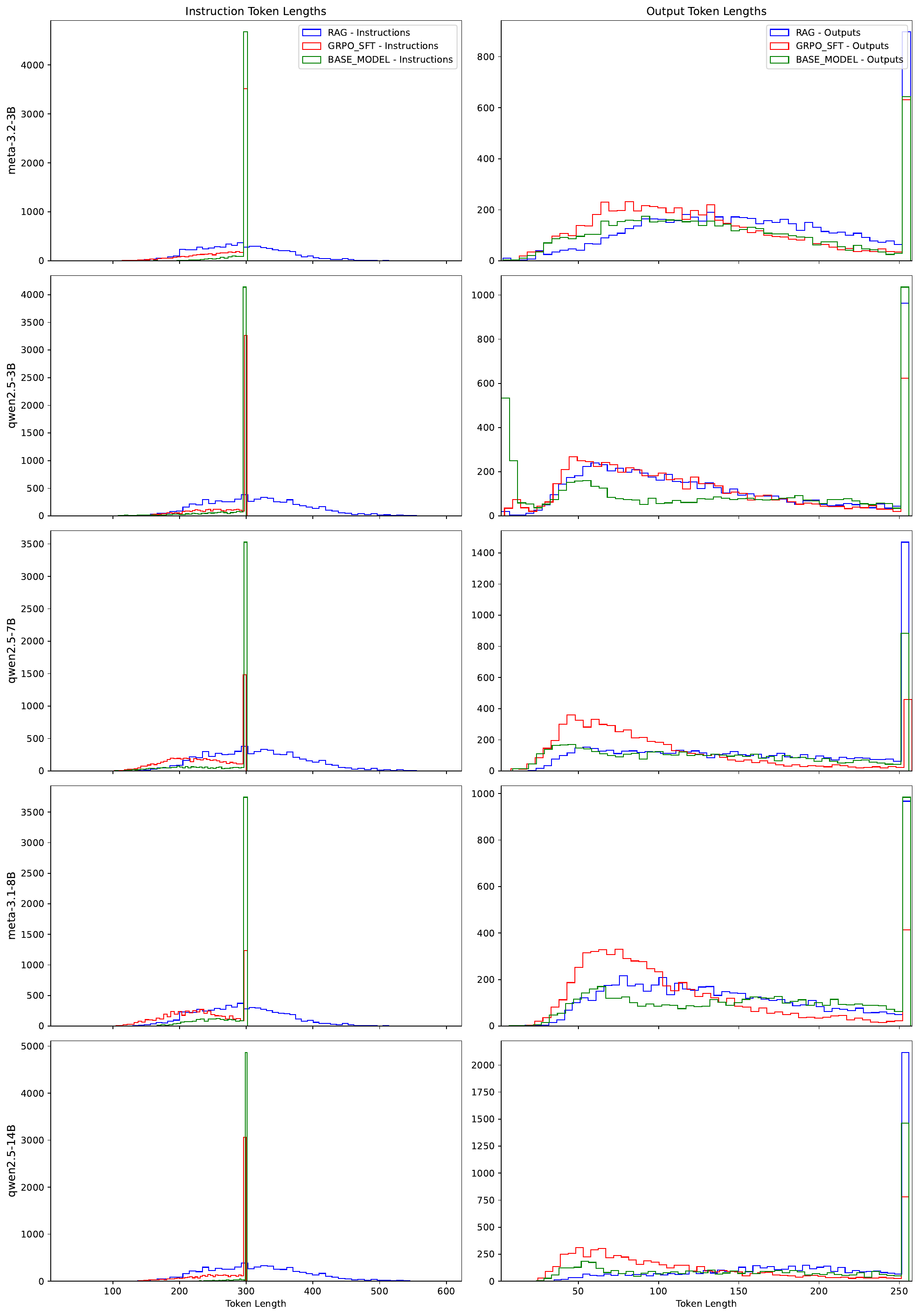}
    \caption{Token distribution length}
     \label{fig:token}
\end{figure*}

Figure~\ref{fig:token} illustrates the token distributions across all models considered. We observe that for the target model outputs, many truncated responses occur with the \texttt{RAG} demonstrations (shown in blue). However, GRAD  generate more concise answers with fewer truncations. This indicates that providing short and focused demonstrations helps the target model to answer concisely and use its limited token budget more effectively for reasoning and final responses.  A similar trend is observed for the demonstration sampler outputs: GRAD produces shorter demonstrations with reduced output length.

Figure~\ref{fig:token} shows the token length distribution for both instructions and outputs across models. Our method, GRAD, is used during both instruction generation and final output generation. Compared to RAG and the base model, GRAD consistently produces shorter prompts and outputs. 

\subsection{Further analysis} \label{apx: further_analysis}

\begin{figure*}[h]
    \centering
    \includegraphics[width=1\linewidth]{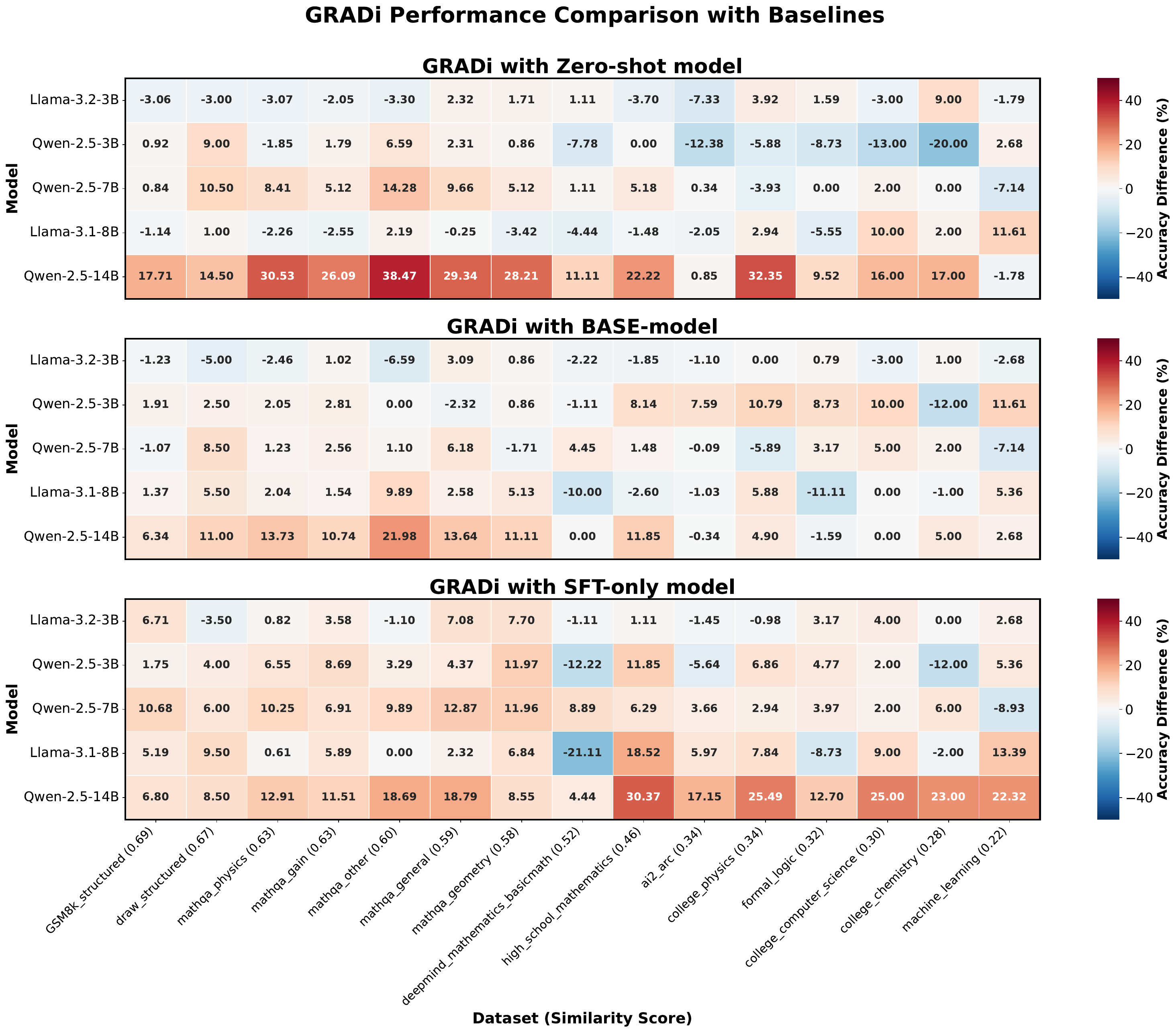}
    \caption{Heatmap of accuracy differences between GRADi and different baselines. Red denotes gains for GRADi, blue for other baselines, respectively, and lighter cells indicate similar performance.}
    \label{fig:grad_vs_baselines}
\end{figure*}

\paragraph{GRADi compared with Zero-shot model.} Figure~\ref{fig:grad_vs_baselines} shows a similar comparison for GRADi versus zero-shot performance. Smaller models on less similar datasets perform better in the zero-shot setup, suggesting they treat GRADi’s demonstrations as noise. In contrast, larger models (Qwen2.5-14B) gain significantly from GRADi on similar datasets, indicating that bigger models leverage demonstrations more effectively than smaller ones.

\paragraph{GRADi compared with BASE model.} Figure~\ref{fig:grad_vs_baselines} compares the accuracy of different target models using the trained GRADi and the untrained (vanilla) model. GRADi works best on inputs similar to the queries but offers smaller gains than ZERO, suggesting BASE produces less relevant demonstrations. It outperforms BASE on smaller models like Qwen2.5-3B, though both lag behind some baselines.

\paragraph{GRADi compared with SFT-only model.} Figure~\ref{fig:grad_vs_baselines} clearly highlights that GRADi outperforms the SFT-only version with most models on nearly all datasets. The most significant improvement is observed with the largest model, particularly when the queries are less similar to the training data.

\newpage
\subsection{\textsc{Prompts}} \label{generationPompt} The model employs two distinct prompts for different stages: one for generating demonstrations and another for producing the final answer. Additionally, we design prompts specific to the benchmark type. For benchmarks that are aligned with our training dataset, we create a default prompt. For the multiple-choice question, we introduce a dedicated prompt to accommodate its format better.

\noindent\textbf{Generating demonstrations:}
\begin{tcolorbox}[promptbox]
You are an educational chatbot specialized in mathematical reasoning. For each question provided by the user, do the following: \\
1. You have a budget of 300 tokens to create **two different examples** that are inspired by the user's input.\\
2. Make sure that the generated examples are short and differ significantly in **context and reasoning steps**.\\
3. Do not repeat the same problem using different numbers or different words. **Each example must be genuinely unique**.\\
4. For each example, describe the problem scenario and its context.\\
5. Provide a step-by-step reasoning to solve the problem defined in the example.\\
6. Finish each example with the numerical answer, labelled clearly as \#\#\#\# [your numerical result here without any units or symbols].\\
7. Append the \texttt{[END\_DEMO]} token at the end of each example.\\
Ensure that the generated examples are short, unique, factually correct, clearly described and solvable. Focus on **maximum educational value**.\\
\end{tcolorbox}

For the multiple-choice question we designed:

\begin{tcolorbox}[promptbox]
You are an educational assistant specializing in multiple-choice science reasoning. For each user question, follow these steps:\\
1. You have a budget of 300 tokens to create **two different multiple-choice examples** that are inspired by the user's input.\\
2. These should test **different scientific ideas** or applications related to the concept.\\
3. Each example must include:\\
- A clearly marked and worded question\\
- Four answer choices labeled A, B, C, and D\\
- A brief explanation of the correct answer\\
- The final answer written as: \#\#\#\# [correct letter]\\
- The \texttt{[END\_DEMO]} token after the final answer.\\
4. Do not copy or rephrase the original question. Focus on generating fresh, diverse examples that are educational and well-structured.\\
5. Remember: Generate **clearly marked** and **unique** questions, answer choices, explanation a final answer and the \texttt{[END\_DEMO]} token.\\
\end{tcolorbox}

\noindent\textbf{Final Answer Generation:}  For final question answering, for the aligned benchmarks the model receives the following system prompt:
\begin{tcolorbox}[promptbox]
You are an educational chatbot specialized in math reasoning.\\
Your task is to solve ONLY the final question provided.\\
DO NOT generate new questions or examples.\\
You are allowed to use only 256 tokens to answer the question.\\
"Provide a step-by-step solution and end with the final numerical answer labeled clearly as \#\#\#\# [your numerical result here without any units or symbols].\\
\end{tcolorbox}

For the multiple-choice question type, the model receives:
\begin{tcolorbox}[promptbox]
Remember:\\
1. Review all choices (A–D)\\
2. Provide a very short explanation using scientific logic\\
3. Do NOT start with the answer\\
4. After the explanation, write only the final answer on a new line like:\\
\#\#\#\# [A single letter A–D]
\end{tcolorbox}
\newpage
\section{Example of demonstrations}

\subsection{ID Example from GSM8K}
This example illustrates the ID setting, where both the task and the retrieved demonstrations come from GSM8K math reasoning problems.

\subsubsection{RAG ID}
\begin{tcolorbox}[promptbox]
Question: Jerry has three times as many stickers as George. George has 6 fewer stickers than his brother Fred. If Fred has 18 stickers, how many stickers does Jerry have?\\

Demonstrations: \\
\#\#\# Question: Dan has two times as many stickers as Tom. Tom has 3 times as many stickers as Bob. If Bob has 12 stickers, how many does Dan have? \\
\#\#\# Answer: First, determine the number of stickers Tom has by multiplying Bob's stickers by 3: 3 x 12 = 36. Next, find Dan's number of stickers by multiplying Tom's stickers by 2: \quad 2 x 36 = 72. Thus, the answer is 72 \texttt{[END\_DEMO]}.\\

\#\#\# Question: Karl, Ryan, and Ben are fond of collecting stickers. Karl has 25 stickers. Ryan has 20 more stickers than Karl. Ben has 10 fewer stickers than Ryan. They placed all their stickers in one sticker book. How many stickers did they place altogether?\\
\#\#\# Answer: To find the total number of stickers, first determine Ryan's count by adding 20 to Karl's 25, which equals 45 stickers. Next, calculate Ben's total by subtracting 10 from Ryan's 45, resulting in 35 stickers. Finally, add Karl's 25, Ryan's 45, and Ben's 35 together: 25 + 45 + 35 = 105 stickers. The answer is 105 \texttt{[END\_DEMO]}.\\
            
Final answer: \\
First, determine the number of stickers George has by subtracting 6 from Fred's 18 stickers: 18 - 6 = 12. Next, find Jerry's number of stickers by multiplying George's stickers by 3: \quad 3 x 12 = 36. Thus, the answer is 36. \#\#\#\# 36 \xmark\\

Correct Answer: 60\\
\end{tcolorbox}

\subsubsection{GRAD ID}
\begin{tcolorbox}[promptbox]
Question: Jerry has three times as many stickers as George. George has 6 fewer stickers than his brother Fred. If Fred has 18 stickers, how many stickers does Jerry have?\\

Demonstrations: \\
Example 1: **Scenario:** A student subscribes to a monthly magazine for 12. Another student buys the same magazine weekly, paying 3 each week. Calculate how much more the second student pays annually.\\
**Reasoning:** \\- Monthly subscription cost: $12 * 12 = 144$\\- Weekly purchase cost: $3 * 52 = 156$ \\- Difference: $156 - 144 = 12$ \#\#\#\# 12 \\ \texttt{[END\_DEMO]} \\ 

Example 2: **Scenario:** Two friends buy coffee every day. One gets a monthly card for 30, while the other buys a cup daily for 2. Find out how much extra the daily buyer spends in a year.\\**Reasoning:**\\- Monthly card cost: $30 * 12 = 360$ \\ - Daily purchase cost: $2 * 365 = 730$ \\- Extra spent: $730 - 360 = 370$\\ \#\#\#\# 370 \texttt{[END\_DEMO]}\\

Final answer: \\
**Reasoning:**\\- Grant's annual cost for newspaper delivery: 200.00\\- Juanita's weekly cost for newspapers: $(0.50 * 6) + 2.00 = 3.00 + 2.00 = 5.00$\\- Juanita's annual cost for newspapers: $5.00 * 52 = 260.00$ \\- Difference in annual costs: $260.00 - 200.00 = 60.00$\\  \#\#\#\# 60 \cmark\\

Correct Answer: 60
\end{tcolorbox}

\subsection{OOD Example from MMLU-college-physics}
We aim to test both RAG and GRAD in an OOD scenario where no in-domain demonstrations are available. The retrieval pool only contains GSM8K math reasoning examples, but the target task is multiple-choice question answering (MCQ) in a completely different domain, namely college physics. This setup allows us to evaluate the generalization ability of the models when faced with mismatched demonstrations. 

\subsubsection{RAG OOD}
For RAG, the retrieved demonstrations are not semantically related to the user query. They are simply the closest available matches, which also makes the reasoning chain unnecessarily long since the demos are not helpful.

\begin{tcolorbox}[promptbox]
Question: \\
What physical property distinguishes elements on the left side of the Periodic Table of the Elements from elements on the right side? \\
A. number of protons \\
B. state of matter\\
C. atomic mass\\
D. temperature\\
                    
Demonstrations: \\
\#\#\# Question: Shawn collected 40 plain pebbles. He painted 9 pebbles red and 13 pebbles blue. He then divided the remaining pebbles equally into 3 groups, and painted them purple, yellow, and green. What is the difference between the number of blue and yellow pebbles? \\  \#\#\# Answer: Shawn collected 40 pebbles in total, painting 9 red and 13 blue. This left 18 pebbles (40 - 9 - 13), which he divided equally into 3 groups of 6. He painted one group yellow, and the difference between blue and yellow pebbles is 13 - 6 = 7. The answer is 7 \texttt{[END\_DEMO]}\\

\#\#\# Question: Megan bought 2 dozen eggs. As she was walking to her car, she dropped a tray of eggs. 3 eggs broke, and twice as many cracked. What is the difference between the eggs that are still in perfect condition and those that are cracked?  \\ \#\#\# Answer: Megan bought 2 dozen eggs, which is 2 x 12 = 24 eggs. When she dropped the tray, 3 eggs broke and twice as many cracked, so 3 x 2 = 6 eggs cracked. The total number of damaged eggs is 3 + 6 = 9 eggs. The difference between the perfect eggs and the cracked eggs is found by subtracting the damaged eggs from the total and then subtracting the cracked eggs: 24 - 9 = 15, and 15 - 6 = 9. The answer is 9. \texttt{[END\_DEMO]} \\

Final answer: \\
Elements on the left side of the Periodic Table tend to be metals, while those on the right side are nonmetals or metalloids. Metals generally have properties such as being good conductors of heat and electricity, having a shiny appearance, and being malleable and ductile. Nonmetals and metalloids do not share these characteristics. However, the key distinguishing factor based on their position is their electron configuration, particularly the number of valence electrons. Elements on the left typically have fewer valence electrons, whereas those on the right have more. This difference in the number of valence electrons affects their chemical behavior and physical properties.\\\\ The number of protons (option A) does not distinguish left-side elements from right-side elements because it defines the element itself rather than its position relative to others.\\\\ State of matter (option B) can vary depending on conditions and does not consistently differentiate left from right.\\\\ Atomic mass (option C) varies widely across the table and does not reliably distinguish left from right.\\\\ Temperature (option D) is an external condition and does not define elemental properties.\\\\ Thus, the correct answer is related to the number of valence electrons, which correlates closely with the number of protons in determining the element's position.\\\#\#\#\# [A] \xmark\\

Correct Answer: B
            
\end{tcolorbox}

\subsubsection{GRAD OOD}

For GRAD, the demonstrations are constructed to be task-aligned and domain-relevant, enabling the model to adapt more effectively to the OOD setting.
\begin{tcolorbox}[promptbox]
Question: \\
What physical property distinguishes elements on the left side of the Periodic Table of the Elements from elements on the right side? \\
A. number of protons \\
B. state of matter\\
C. atomic mass\\
D. temperature\\
                         
Demonstrations: \\
\#\#\# Example 1:**Question:** Which characteristic is more likely to be found in metals located on the left side of the periodic table compared to non-metals on the right side?\\\\A. They are typically gases at room temperature  \\B. They tend to lose electrons easily  \\C. They have lower melting points  \\D. They form acidic oxides  \\\\**Explanation:** Metals on the left side of the periodic table tend to lose electrons easily, making them good conductors of electricity and heat. Non-metals on the right side often gain electrons, forming negative ions. This contrasts with the metallic behavior seen on the left.\\\\\#\#\#\# [correct letter]: B \texttt{[END\_DEMO]}\\\\

\#\#\# Example 2:**Question:** How does the typical reactivity of elements change as you move from the left to the right across the periodic table?\\\\A. Reactivity increases because the atomic radius decreases  \\B. Reactivity decreases because the atomic mass increases  \\C. Reactivity remains constant due to similar electron configurations  \\D. Reactivity increases due to higher temperatures  \\\\**Explanation:** As you move from left to right across the periodic table, the atomic radius generally decreases, leading to increased reactivity for metals (left) and decreased reactivity for non-metals (right), due to their ability to attract electrons.\\\\\#\#\#\# [correct letter]: A \texttt{[END\_DEMO]}\\\\
            
Final answer: \\
**Explanation:** The state of matter differs significantly between elements on the left and right sides of the periodic table. Elements on the left are predominantly metals, which are solid at room temperature, while elements on the right include non-metals that can exist as gases or liquids at room temperature.\\\\\#\#\#\# [correct letter]: B \cmark \\

Correct Answer: B 
\end{tcolorbox}

%% ORIGINAL TABLE
\section{Extended Results}
We present all results across the six different strategies using models of varying sizes. In all experiments, the demonstration sampler and the target model are the same.  We investigate both small- and medium-scale models. For small-scale settings, we evaluate with 3B parameter models such as \textbf{LLaMA3.2-3B Instruct} and \textbf{Qwen2.5-3B Instruct}. For medium- to large-scale settings, we include \textbf{Qwen2.5-7B Instruct}, \textbf{LLaMA3.1--8B Instruct}, and \textbf{Qwen2.5-14B Instruct}. 

Table~\ref{apx:performance_cleaned} presents all results. We observe that our method does not perform well only with the 3B architectures. For \textbf{LLaMA3.2-3B}, the zero-shot setting dominates, indicating that the model can answer questions effectively without relying on additional context. Since the model is relatively small, even relevant context can introduce confusion and reduce performance. 

In contrast, for \textbf{Qwen 3B}, the RAG strategy performs best, suggesting that Qwen benefits from demonstrations. However, our method still underperforms with these lightweight models, which indicates that the demonstration sampler may not be adequate for such small model sizes. Overall, 3B models appear insufficient for generating high-quality demonstrations compared to their larger counterparts.

\begin{table*}[h]
    \centering
    \renewcommand{\arraystretch}{1}
    \begin{tabular}{l | l | c c c c c c c}
        \hline
        \multirow{2}{*}{\textbf{Model}} & \multirow{2}{*}{\textbf{Method}} & \multicolumn{6}{c}{\textbf{Dataset}} \\
        \cline{3-8}
        & & \textbf{GSM8K} & \textbf{\makecell{draw\\structured}} & \textbf{MathQA*} & \textbf{\makecell{deepmind\\basic\_math}} & \textbf{\makecell{ARC\\Challenge}} & \textbf{MMLU*} \\
        \hline

        \multirow{6}{*}{\makecell{LLaMA\\3.2-3B\\Instruct}} 
        & Zero-shot  & \textbf{71.91} & 33.00 & \textbf{39.86} & 38.89 & \textbf{71.84} & 38.02  \\
        & RAG       & 67.48 & \textbf{37.50} & 38.74 & 38.89 & 27.82 & 15.19  \\
        & SFT-only   & 62.14 & 33.50  & 35.20 &   41.11    &    65.96       & 36.42 \\
        & BASE       & 70.08 & 35.00 & 38.95 & 42.22 & 65.61 & \textbf{39.14}\\
        & \bluecell{GRAD}   &  68.70     &  32.50     &  38.74     &   \bluecell{43.33}       &   64.25       & 38.64 \\
        & \greencell{GRADi}       & 68.85 & 30.00 & 39.54 & 40.00 & 64.51 & 38.02  \\
        \hline

        \multirow{6}{*}{\makecell{Qwen\\2.5-3B\\Instruct}} 
        & Zero-shot  & 71.98 & 30.00 & 43.29 & 47.78 & \textbf{76.71} & \textbf{50.25}  \\
        & RAG       & \textbf{74.73 }& \textbf{39.00} & \textbf{46.51} & \textbf{52.22} & 30.89 & 21.36  \\
        & SFT-only   & 71.15  & 35.00 & 38.26 & \textbf{52.22} & 69.97 & 39.38 \\
        & BASE      & 70.99 & 36.50 & 44.31 & 41.11 & 56.74 & 37.66  \\
        & \bluecell{GRAD}   & 70.31 & 38.50 & 44.96 & 43.33 & 67.58 & 43.09 \\
        & \greencell{GRADi}       & 72.90 & \greencell{39.00} & 44.52 & 40.00 & 64.33 & 44.44  \\
        \hline

        \multirow{6}{*}{\makecell{Qwen2.5\\7B\\Instruct}} 
        & Zero-shot  & 83.89 & 36.50 & 44.79 & 67.78 & 87.71 & 62.22  \\
        & RAG        & 83.59 & 36.50 & 43.78 & 63.33 & 85.92 & 59.51  \\
        & SFT-only   & 74.05 & 41.00 & 42.38 & 60.00 & 84.39 & 59.88 \\
        & BASE       & \textbfc{85.80} & 38.50 & 49.73 & 64.44 & \textbfc{88.14} & 62.59  \\
        & \bluecell{GRAD}   & 84.27 & \bluecell{43.00} & \bluecell{54.72} & \bluecell{70.00} & 88.05 & \bluecell{64.20} \\
        & \greencell{GRADi}       & 84.73 & \greencell{47.00} & \greencell{53.11} & \greencell{68.89} & 88.05 & \greencell{62.71}  \\
        \hline

        \multirow{6}{*}{\makecell{LLaMA\\3.1-8B\\Instruct}} 
        & Zero-shot  & 78.24 & 42.00 & 44.04 & 43.33 & \textbfc{83.53} & 49.13  \\
        & RAG        & 76.79 & 29.50 & 40.67 & 38.89 & 73.89 & 39.75  \\
        & SFT-only   & 71.91 & 33.50 & 39.91 & \textbfc{60.00} & 75.51 & 42.72 \\
        & BASE       & 75.73 & 37.50 & 39.97 & 48.89 & 82.51 & \textbfc{52.47}  \\
        & \bluecell{GRAD}   & \bluecell{78.85} & \bluecell{46.50} & \bluecell{45.12} & 46.67 & 80.80 & 50.00 \\
        & \greencell{GRADi}       & 77.10 & \greencell{43.00} & 42.70 & 38.89 & 81.48 & 51.23  \\
        \hline

        \multirow{6}{*}{\makecell{Qwen2.5\\14B\\Instruct}} 
        & Zero-shot  & 72.75 & 30.50  & 27.73 & 58.89 & 91.13 & 48.27  \\
        & RAG        & 83.89 & 27.50 & 37.50 & 64.44 & 90.70 & 48.52  \\
        & SFT-only   & 83.66 & 36.50 & 42.00 & 65.56 & 74.83 & 40.74 \\
        & BASE       & 84.12 & 34.00 & 43.78 & 70.00 & \textbfc{92.32} & 59.75  \\
        & \bluecell{GRAD}    & \bluecell{90.92} & \bluecell{40.50} & \bluecell{56.98} & \bluecell{72.22} & 91.64 & \bluecell{65.31} \\
        & \greencell{GRADi}       & \greencell{90.46}  &\greencell{45.00} & \greencell{57.80} & \greencell{70.00} & 91.98 & \greencell{65.06}  \\
        \hline
    \end{tabular}
    \caption{Performance Comparison across Models and Methods (Accuracy in \%). The same backbone model is used for both the demonstration sampler and the target model. Datasets are ordered by their semantic similarity from left to right in decreasing order. Blue indicates cases where GRAD outperforms all baselines (independent of GRADi), while green indicates cases where GRADi does so (independent of GRAD). If the best-performing model on a given benchmark is neither GRAD nor GRADi, it is reported in bold.}
    \label{apx:performance_cleaned}
\end{table*}
\end{document}